%% file: SmartPaw_ICAR2023.tex
\documentclass[letterpaper, 10 pt, conference]{ieeeconf}  

\IEEEoverridecommandlockouts                              

\usepackage{graphics} 
\usepackage{svg}
\usepackage{graphicx}
\usepackage{subcaption}

\usepackage{epsfig} 
\usepackage{amsmath} 
\usepackage{amssymb}  
 \usepackage{algorithm}
\usepackage[compatible]{algpseudocode}
\algnewcommand\AAND{\textbf{ and }}
\algnewcommand\Or{\textbf{ or }}
\usepackage{citesort}
\usepackage{flushend}
\usepackage{url}

\usepackage[colorinlistoftodos]{todonotes}

\DeclareMathAlphabet{\pazocal}{OMS}{zplm}{m}{n}

\newcommand{\ignore}[1]{}

\DeclareMathAlphabet{\mathpzc}{OT1}{pzc}{m}{it}

\definecolor{LightCyan}{rgb}{0.5,1,0.5}

\usepackage{array}
\newcolumntype{C}[1]{>{\centering\arraybackslash}p{#1}}
\newcolumntype{M}[1]{>{\raggedright\arraybackslash}p{#1}}

\usepackage{array} 
\newcolumntype{L}[1]{>{\raggedright\let\newline\\\arraybackslash\hspace{0pt}}m{#1}}	
\newcolumntype{S}[1]{>{\centering\let\newline\\\arraybackslash\hspace{0pt}}m{#1}}
\newcolumntype{R}[1]{>{\raggedleft\let\newline\\\arraybackslash\hspace{0pt}}m{#1}}

\usepackage[nolist,nohyperlinks]{acronym}
\input{acronyms.tex}

\makeatletter
\renewcommand*{\@opargbegintheorem}[3]{\trivlist
  \item[\hskip \labelsep{\itshape #1\ #2}] \textit{(#3)}\ }
\makeatother

\title{\LARGE \bf
Terrain Recognition and Contact Force Estimation through a Sensorized Paw for Legged Robots
 }

\author{ 
Aleksander Vangen$^1$, Tejal Barnwal$^2$, J{\o}rgen Anker Olsen$^1$, and Kostas Alexis$^1$
\thanks{
}%
\thanks{$^1$The authors are with Norwegian University of Science and Technology (NTNU), O.S. Bragstads Plass 2D , 7034, Trondheim, NO {\tt\small jorgen.a.olsen@ntnu.no}}
\thanks{$^{2}$The author is with Indian Institute of Technology Bombay, India {\tt\small tejalbarnwal.iitb23@gmail.com}}
\newline
}

\begin{document}

\maketitle
\thispagestyle{empty}
\pagestyle{empty}

\input{000Abstract}

\section{INTRODUCTION}\label{sec:intro}
\input{010_Introduction}

\section{RELATED WORK}\label{sec:related}
\input{020_RelatedWork}

\section{SYSTEM DESIGN}\label{sec:sys_des}
\input{030_SystemDesign.tex}

\section{DATA COLLECTION}\label{sec:data_col}
\input{040_DataCollection.tex}

\section{EXPERIMENTAL EVALUATION}\label{sec:experiments}

\input{050_Experiments}

\section{CONCLUSIONS}\label{sec:conclusions}

\input{060_Conclusion}

\bibliographystyle{IEEEtran}
\bibliography{./BIB/SmartPaw_ICAR2023.bib}

\end{document}

%% file: acronyms.tex
\acrodef{mlt}[MLTs]{Martian Lava Tubes}
 
\acrodef{smartpaw}[T\scriptsize {RACE}\normalsize Paw]{Terrain Recognition And Contact force Estimation Paw}

\acrodef{mae}[MAE]{Mean Absolute Error}

%% file: 000Abstract.tex
\begin{abstract}
This paper introduces the Terrain Recognition And Contact Force Estimation Paw, a compact and sensorized shoe designed for legged robots. The paw end-effector is made of silicon that deforms upon the application of contact forces, while an embedded micro camera is utilized to capture images of the deformed inner surface inside the shoe, and a microphone picks up audio signals. Processed through machine learning techniques, the images are mapped to compute an accurate estimate of the cumulative $3$D force vector, while the audio signals are analyzed to identify the terrain class (e.g., gravel, snow). By leveraging its on-edge computation ability, the paw enhances the capabilities of legged robots by providing key information in real-time that can be used to adapt locomotion control strategies. To assess the performance of this novel sensorized paw, we conducted experiments on the data collected through a specially-designed testbed for force estimation, as well as data from recordings of the audio signatures of different terrains interacting with the paw. The results demonstrate the accuracy and effectiveness of the system, highlighting its potential for improving legged robot performance.

\end{abstract}

%% file: 010_Introduction.tex
Autonomous robots capable of traversing challenging terrain are necessary for a host of critical applications. Legged robot systems, in particular, have garnered significant attention due to their applicability in diverse fields such as industrial inspection, search and rescue, and extraterrestrial exploration~\cite{kim2017design,agha2021nebula,tranzatto2022cerberus,tranzatto2022team}. The efficiency, robustness, and safety of these systems hinge on their capability to accurately estimate the contact forces between the robot's foot and the ground. Moreover, the detection of the underlying terrain type (e.g., concrete, gravel, snow) is an essential factor that can augment their capabilities, facilitating more robust and adaptive locomotion. Consequently, this enables legged robots to navigate through a wide range of challenging environments with increased robustness and efficiency.

\begin{figure}
    \centering
    \includegraphics[width = 0.99\columnwidth]{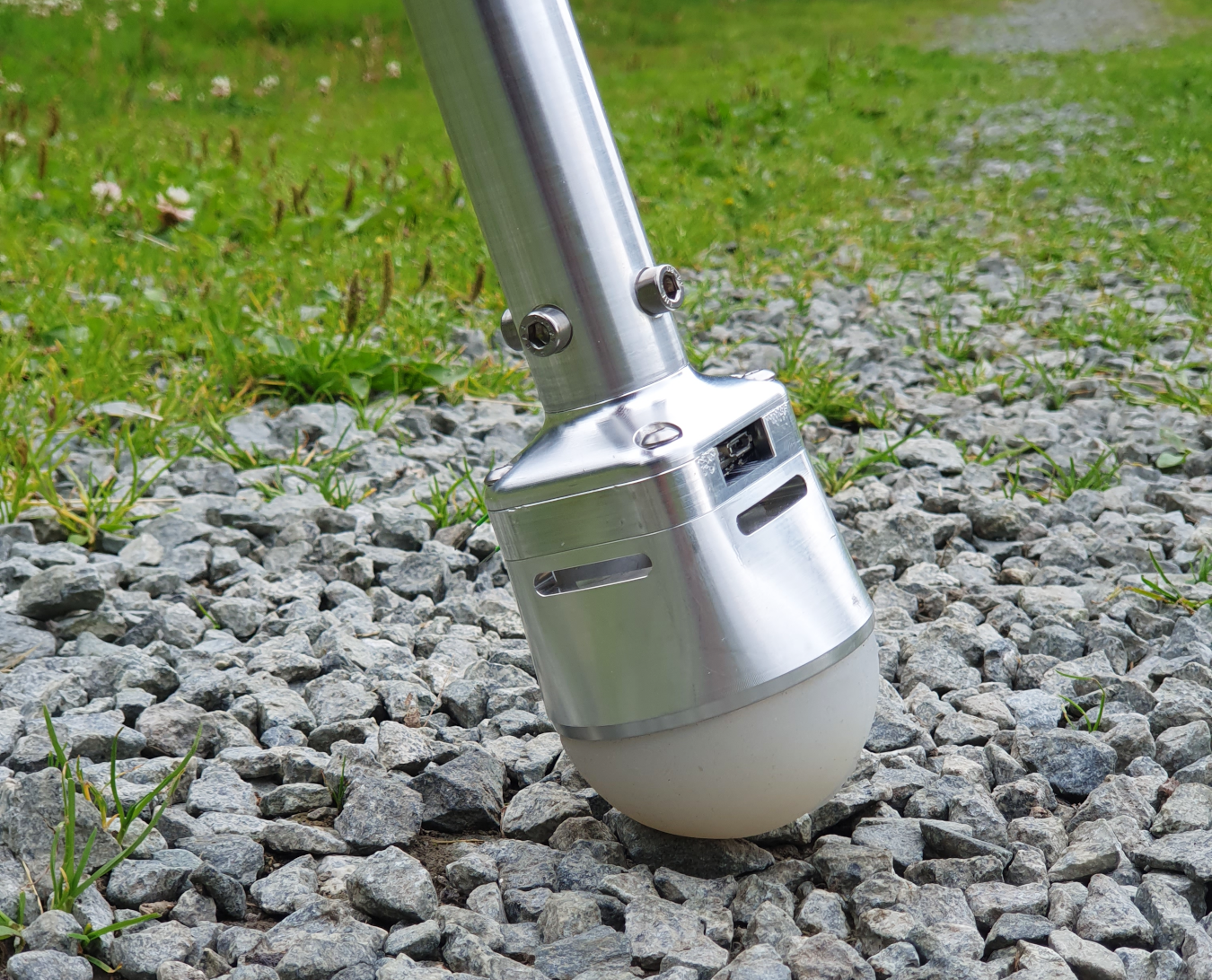}
    \vspace{-3ex}
    \setlength{\belowcaptionskip}{-20pt}
    \caption{The developed \ac{smartpaw} system delivering contact force estimation and terrain classification on the edge for legged robots.}
    \label{fig:introfigure}
    \vspace{-1ex}
\end{figure}

While prior research has contributed to the advancement of shoe sensorization \cite{kolvenbach2020towards,kolvenbach2019haptic,kolvenbach2019tactile,hauser2016friction}, the scope of these studies has been largely focused on niche applications, such as planetary soil sampling, leading to domain-specialized shoe designs. However, a look at successful legged robots indicates that conventional ``point-foot’’ solutions are the most widely used type owing to their structural simplicity and robustness. Motivated by the above, this study introduces the \ac{smartpaw}, an innovative sensorized point-foot shoe for legged robots that exploits embedded vision and audio signal processing to deliver contact force estimation on the paw and soil type characterization. The final system is depicted in Figure~\ref{fig:introfigure}.

The \ac{smartpaw} incorporates a compliant hemispherical contact point that on its inner side presents a pattern that is captured by an embedded micro camera positioned within the shoe itself, alongside a microphone and the necessary compute. Exploiting this design, it embodies four pivotal attributes. First, it consolidates image processing and supervised learning techniques in order to estimate the cumulative contact force and the direction of its exertion, while effectively circumventing the need for complicated and typically parametrically uncertain modeling of material deformation properties. Secondly, it incorporates an advanced terrain classification scheme relying on a neural network processing the unique acoustical signatures generated during the interaction between the paw and the terrain. Thirdly, the \ac{smartpaw} has been designed with edge processing in mind, meaning that all the presented computations take place within the shoe onboard a power-efficient microcontroller thus not requiring complex wiring within the robot’s body. Finally, this sensorized shoe employs a pragmatic approach to design, incorporating commercially off-the-shelf electronics, and boasting ease of fabrication, thereby ensuring accessibility and scalability in its application.

To validate the proposed design, a series of experimental studies were conducted. These studies involve force estimation using the \ac{smartpaw} under various load conditions and orientations, as well as terrain classification using a variety of terrain surfaces and soil types. The results were obtained using experiments based on a specialized testing set-up allowing to measure forces alongside labeled data for the terrain class. The results highlight the accuracy of force estimation and terrain classification provided by the system. To allow community reuse and reproducibility, we release the solution as an open-hardware and open-software project at \url{https://github.com/ntnu-arl/trace_paw}. 

The remainder paper is organized as follows: Section \ref{sec:related} overviews related work in shoe design for legged robots. Section~\ref{sec:sys_des} details the mechanical design of the \ac{smartpaw}, in addition to describing the methods used for vision-based force estimation and audio-based terrain classification. The experimental setup is described in Section~\ref{sec:data_col}, while Section~\ref{sec:experiments} outlines the results obtained. The conclusions drawn are subsequently discussed in Section~\ref{sec:conclusions}.

%% file: 020_RelatedWork.tex
Three niche directions of research, namely that of a) ``shoe'' design for legged robots, b) vision-based force estimation, and c) audio-based terrain classification relate to the contribution in this work. In the domain of shoe end effector design, recent work has focused on developing specialized feet for legged robots such as quadrupeds. Generally, the most common shoe design is that of ``point foot'' as found in successful legged robot designs such as ANYmal~\cite{hutter2016anymal} and Spot~\cite{guizzo2019leaps}. Passive point feet offer a flexible solution across most terrains but more sophisticated options are possible both in terms of sensorization and in terms of mechanical design. To that end, the authors in~\cite{kaslin2018towards} present a lightweight, passive, adaptive foot design that implements three rotational degrees of freedom with the goal of offering superior traction on diverse terrains. The works in~\cite{catalano2021adaptive,pollayilsoftfoot} outline the SoftFoot-Q articulated foot made to exhibit reduced slippage and enhanced stability. Inspired by the presence of soft-pads on animals, the contribution in~\cite{hauser2016friction} offers a shoe that allows for high traction and strong damping. Considering the benefits of sensorized feet, the efforts in~\cite{kolvenbach2020towards,kolvenbach2019tactile,kolvenbach2019haptic} present systems that use force and inertial sensing allowing to assess terrain conditions. The involved designs are particularly application-specific with~\cite{kolvenbach2020towards,kolvenbach2019tactile} focusing on assessing concrete deterioration in sewers, and~\cite{kolvenbach2019haptic} targeting the haptic inspection of planetary soils. Focusing on walking on soft terrain, the publication in~\cite{dang2019research} discusses the research status on foot design for legged robots walking on soft terrain. In addition, a host of designs consider diverse challenging tasks for shoe design such as ball kicking~\cite{schempf1995roboleg} and interacting with snowfields~\cite{hakamada2019retrofit}, while the works in~\cite{minakata2004study,minakata2008study} consider the benefit of flexible shoe designs to increase the efficiency of walking robots. In relation to this literature, the proposed system contributes in the area of the widely used and mechanically robust point feet designs albeit with advanced sensorization and the ability to infer both the forces from the environment and the type of terrain with which a robot may interact. 

To that end, vision-based tactile sensing and force estimation literature is particularly relevant. The body of works in~\cite{sferrazza2019ground,sferrazza2020learning,sferrazza2022sim,trueeb2020towards} offer a detailed overview of a system involving multi-camera based visual inference of force relying on neural networks and possessing the potential for utilization as an approach to robotic skin. The authors in~\cite{zhang2018robot} focus on a stereo vision for estimating the contact force field. The contribution in~\cite{yamaguchi2019recent} surveys recent literature on tactile sensing for robotic manipulation with a focus on vision integration. From a different standpoint, the effort in~\cite{xie2012pixel} investigates the use of a novel fiber optic sensor to measure normal forces. Exploiting velocimetry for multi-modal object detection and force feedback, the authors in~\cite{wang2021novel} offer another solution to tactile sensing. Tailored to thumb-sized sensing, the work in~\cite{sun2022soft} presents a technique to derive a directional force-distribution map over its sensing surface and also relies on neural networks. Using stereo vision, the authors in~\cite{zhang2022tac3d} focus on estimating the friction coefficient. Furthermore,~\cite{zhang2022deltact} proposes a new design on vision-based tactile sensing exploiting a dense color pattern and delivers low error and high measurement frequency. Our work builds further upon such ideas and focuses on miniaturized ``edge'' implementation inside the robot's shoe/paw, alongside delivering multi-modality to not only estimate force but infer the terrain/soil type. 

In the domain of audio-based classification, a host of works exist using both conventional and deep learning-based techniques~\cite{hershey2017cnn,guo2003content}. The developed methods and systems have found applications in robotics. Examples include the work in~\cite{romano2013ros} offering end-to-end tools for supervised learning of audio events, feature extraction, classification using support vector machines, alongside detection of audio events. Of relevance to this work are contributions such as those in~\cite{valada2018deep,zurn2020self,valada2017deep} delivering audio-based terrain classification and acoustic feature learning including with visual self-supervision. Furthermore, the authors in~\cite{christie2016acoustics} offer a method for audio-based classification of terrain for legged robots using support vector machines. Our work extends such ideas and focuses on edge implementation within the paw itself.

%% file: 030_SystemDesign.tex
This section introduces the design of the \ac{smartpaw}.

\subsection{Mechatronic Design}
\begin{figure}
     \centering
         \begin{subfigure}[b]{0.19\textwidth}
             \centering
             \includegraphics[width=0.99\columnwidth]{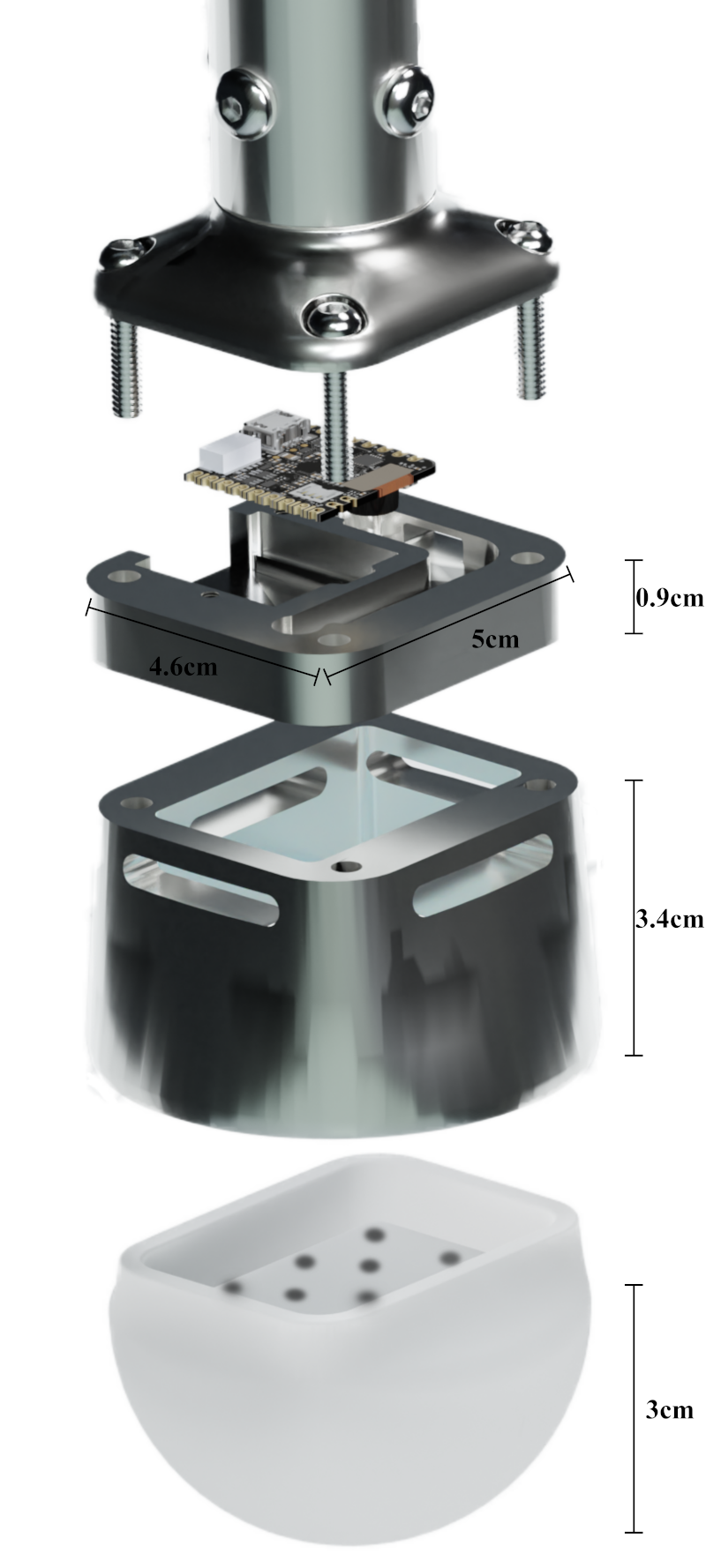}
             \caption{Exploded View}
             \label{fig:exploded_view}
         \end{subfigure}
         \hfill
         \begin{subfigure}[b]{0.262\textwidth}
            \begin{subfigure}[b]{1.0\textwidth}
                 \centering
                 \includegraphics[width=\textwidth]{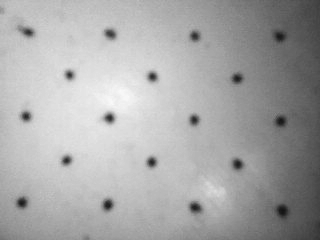}
                 \caption{Undistorted Silicone Surface}
                 \label{fig:undistorted_pattern}
             \end{subfigure}
             \vfill
            \begin{subfigure}[b]{1.0\textwidth}
                 \centering
                 \includegraphics[width=\textwidth]{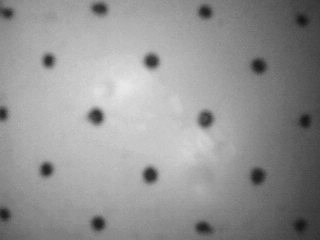}
                 \caption{Distorted Silicone Surface}
                 \label{fig:distorted_pattern}
            \end{subfigure}
         \end{subfigure}
    \setlength{\belowcaptionskip}{-20pt}
    \caption{(\protect\subref{fig:exploded_view}) from top to bottom; shank adapter, Nicla Vision, electronics housing, structural frame \& silicone sole of the \ac{smartpaw}. (\protect\subref{fig:undistorted_pattern}) \& (\protect\subref{fig:distorted_pattern}) silicone patterns in the absence \& presence of deformation-causing forces on the sole.}
    \label{fig:cad_and_pattern}
\end{figure}

Figure~\ref{fig:exploded_view} illustrates the \ac{smartpaw}, which comprises a) the mechatronic design, including the sensing and computing unit, an aluminum frame, and a soft-material-based sole incorporating a grid-like pattern inscribed on its inner surface, as well as b) the associated software to realize the desired force and terrain inference results.  

The paw of the \ac{smartpaw} is specifically designed to be roughly proportionate in size to the shoe size of the ANYmal legged robot~\cite{hutter2016anymal}. This design choice ensures compatibility between the paw and that robot's foot structure while generally rendering size-wise compatible with most established quadrupeds, including ANYbotics ANYmal, Boston Dynamics Spot, Unitree B1, and more.

The design leverages the Arduino Nicla Vision, which integrates a Dual ARM Cortex M7/M4 with a GC2145 camera (2MP), and an MP34DT06JTR microphone onto a single electronic board (dimensions $22.86\times22.86\times 1\textrm{mm}$), thereby simplifying the circuit and facilitating a compact shoe design with all its processing embedded into it. The images captured by the camera and the audio picked up by the microphone are fed into their respective neural networks for force estimation and terrain recognition. Two small white LEDs are mounted next to the Nicla Vision board, pointing downwards to ensure uniform illumination within the shoe.

The compliant sole of the shoe has a solid hemispherical bottom touching the ground that transitions into a flat rectangular surface at the top, where its visual pattern is engraved. The sole is vacuum-cast using silicone of Shore Hardness 10A, and the marked pattern consisting of triangularly arranged points is coated with dark-colored silicone-based ink to introduce contrast. Upon application of contact forces, the soft silicone sole undergoes deformation, causing the marked points to displace and expand, resulting in distinct patterns captured in the images (see Figure~\ref{fig:undistorted_pattern} and \ref{fig:distorted_pattern}). Therefore, these images provide a visual representation of the deformation characteristics of the distorted material surface, thereby serving as an input to derive a map to contact forces.

\subsection{Vision-based 3D Force Estimation}
The force vector inference problem involves correlating the images of the silicone upper surface with corresponding cumulative contact forces. While traditional constitutive models attempt to capture the complex hyperelastic behavior of silicone, they often incorporate simplifying modeling assumptions, such as, the nature of loads \cite{dydo1995elasticity}, which can introduce imperfections. For instance, assuming a uniform load distribution when modeling the soft silicone sole may overlook non-uniform pressure distribution in real-life scenarios. 
These approximations can lead to inaccurate predictions, especially when the paw is subjected to diverse terrain and a wide range of linear and non-linear loading conditions. These models solely rely on point displacement derived from the image, disregarding other indirect features like slight variations in marker point size and the effect of light reflections on the inner silicone surface. In addition, these models neglect manufacturing defects that may impact the intrinsic properties of silicone. To address these limitations, a neural network utilizing pattern-based images is employed. The pattern captured in the image provides good contrast (assisted by the pattern coloring and the incorporated illuminating LEDs within the shoe), which helps the neural network to infer the force vector by identifying the relative direction and extent of point movement along with other secondary factors.

The Neural Network (NN) architecture requires substantial training data for accurate force prediction. For this purpose, grayscale HQVGA ($240\times160$ pixels) raw images streamed through Nicla Vision served as input features, while the corresponding ground truth labels include the three-component force measurements obtained using an ATI Mini45 Force/Torque (F/T) sensor. An external camera captures Aruco tags attached to the paw and the F/T sensor setup to determine the relative orientation between the two. Section \ref{sec:data_col} provides a comprehensive description of the data acquisition process, which involved the integration of three sub-setups within a test rig, along with information regarding the dataset and its associated attributes.

\begin{figure}
     \centering
    \includegraphics[scale=0.04]{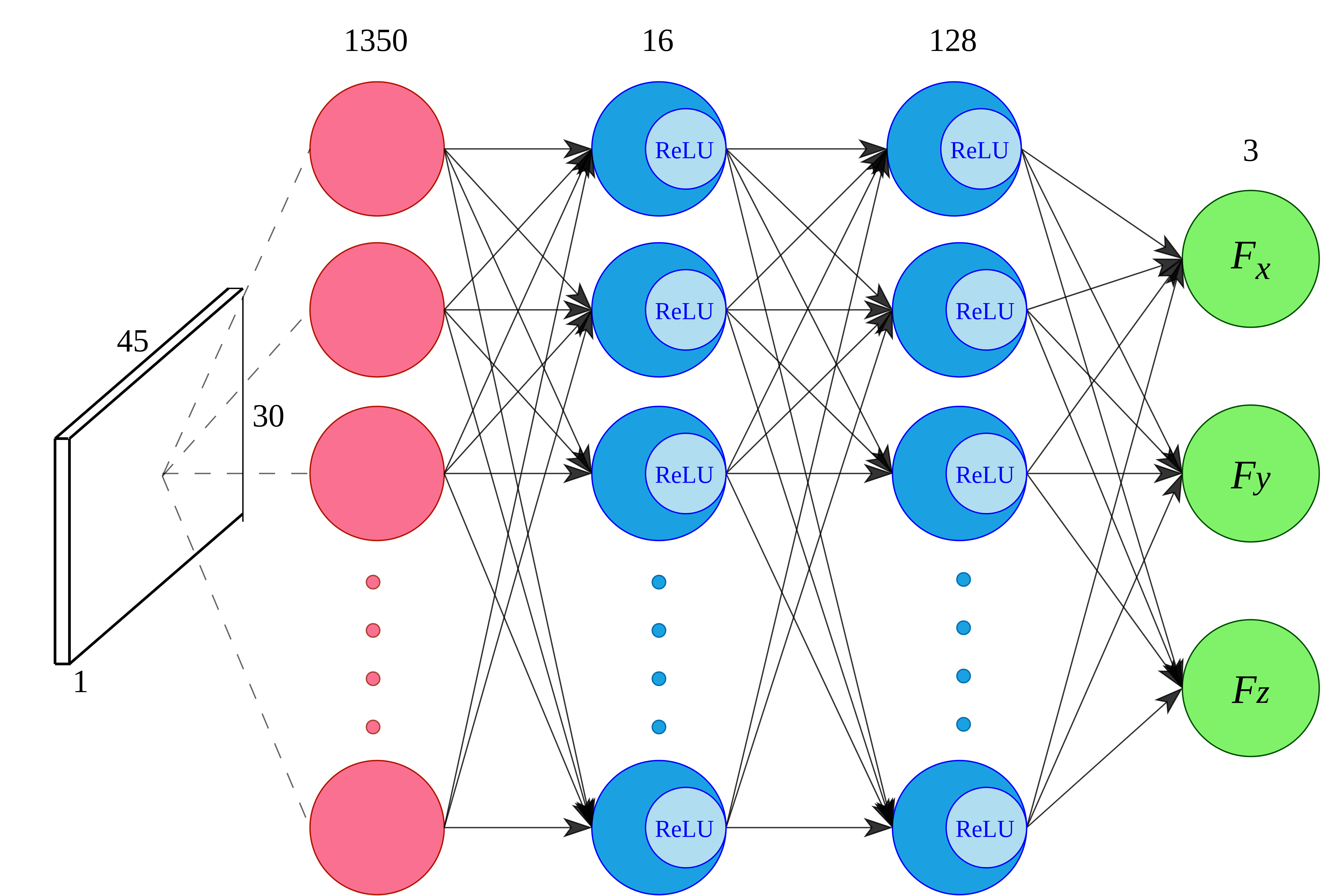}         
    \setlength{\belowcaptionskip}{-20pt}
    \caption{The NN architecture for force estimation. In white, the input layer; in red, the flattening layer; in blue, two hidden layers with $16$ and $128$ neurons, respectively, using ReLU activation function; in green, the output layer with linear activation function to perform the regression task.}
    \label{fig:force_estimation_nn_diagram}
\end{figure}

The neural network training involved adopting a relatively small, Fully Connected Neural Network (FCNN) architecture tailored to accommodate the memory limitations of Nicla Vision. The architecture consisted of an input layer processing resized $45\times 30$ images subsequently flattened, followed by two hidden layers utilizing the Rectified Linear Unit (ReLU) activation function to capture nonlinear relationships within the data. The output layer yielded normalized 3D force vectors, addressing the multivariate regression task. 
To optimize the model's performance, a grid search was conducted to explore various hyperparameters, including the number of hidden layers, number of neurons per layer, learning rate, and dropout. Batch normalization was applied across all the layers together with output scaling to expedite network training. The training employed \ac{mae} as the loss function for the Adam optimizer and incorporated L2 regularization to mitigate over-fitting. By systematically training and evaluating the model for each combination of hyperparameters, the most effective configuration was identified based on performance metrics assessed on the validation set. A diagram summarizing the chosen architecture is shown in Figure~\ref{fig:force_estimation_nn_diagram}.

\subsection{Audio-based Terrain Classification}

Beyond force estimation, this work aims to equip legged robots with the ability to infer the type of terrain they interact with. Classifying terrain based on contact sounds is a task that can be solved by supervised learning, as the problem relies on categorizing the impact-sound data based on distinguishable audio features. The terrain classes under consideration encompass gravel, concrete, leaves, snow, sand, and grass. To extract the frequency data from the audio signal efficiently, the Mel-frequency cepstral coefficients (MFCCs)~\cite{logan2000mel} were utilized and used as the basis to classify the terrain. MFCCs collectively make up the mel-frequency cepstrum of a signal which represents its short-term power spectrum.

The dataset collection involved activating the 16 kHz omnidirectional microphone on the Nicla Vision for one-second intervals while deliberately striking the paw into the ground to emulate the ground interactions of a walking quadrupedal robot. For each terrain class, the collected audio samples were truncated to a length of 62.5 ms to slice out the unimportant background noise and focus on the impact data. The MFCCs were found using a Short-Time Fourier Transform (STFT) with a frame size of 512 bytes and a hop size of 160 bytes. To calculate the mel energies, twenty mel-filter banks were utilized~\cite{huang2001spoken}. Finally, 13 MFCCs were extracted using the Discrete Cosine Transform (DCT), which would act as inputs to the machine learning techniques. 

\begin{figure}
     \centering
    \includegraphics[scale=0.04]{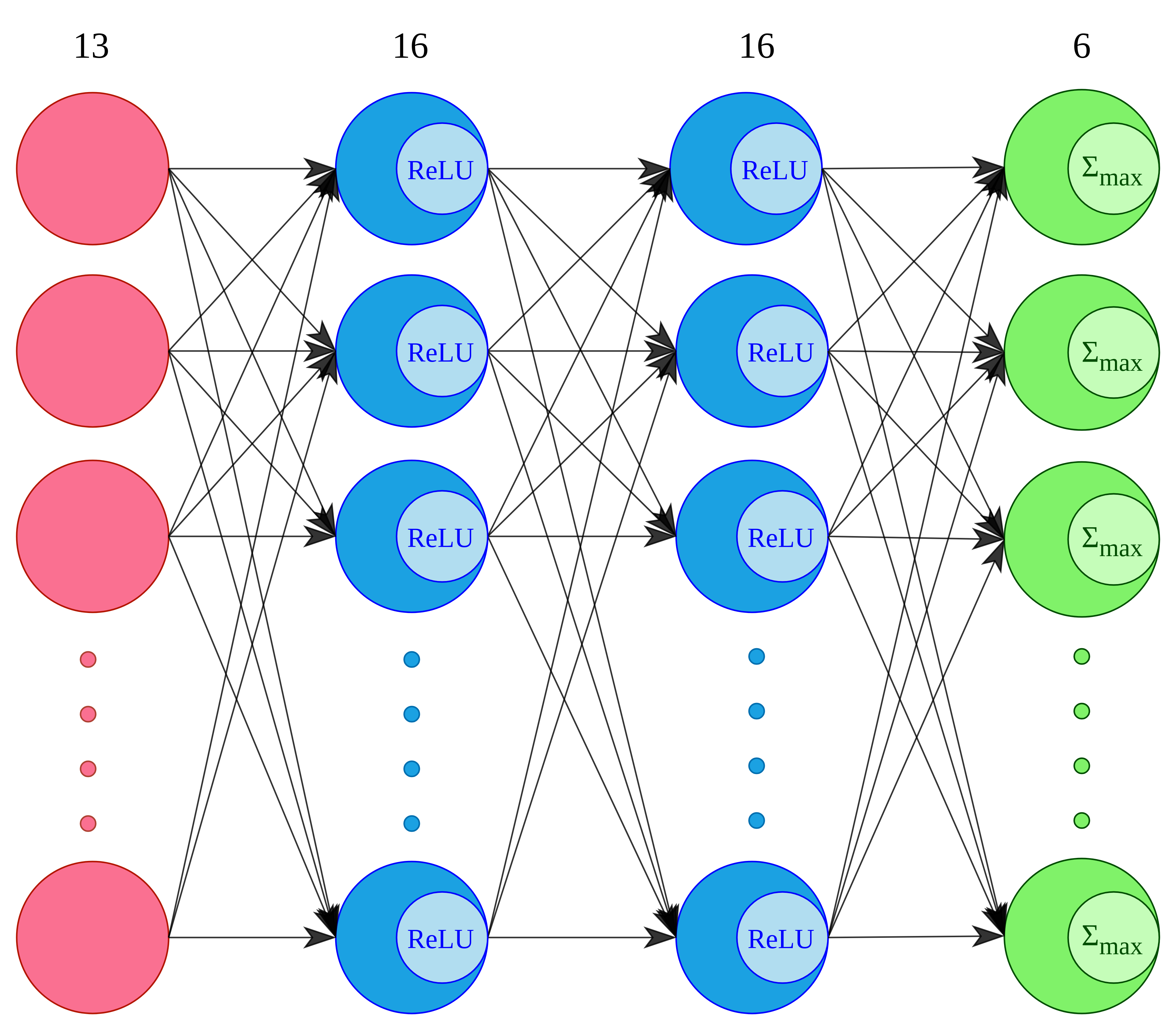}
    \setlength{\belowcaptionskip}{-20pt}
    \caption{The NN architecture for audio-based terrain classification. In red, the input layer with a $13$-element MFCC-based feature vector; in blue, two hidden layers using ReLU activation function; in green, the output layer with softmax activation yielding six terrain class probabilities.}
    \label{fig:terrain_recognition_nn_diagram}
\end{figure}

Owing to the limited memory capacity of the Nicla Vision, lightweight machine learning approaches, including Support Vector Machines, K-Nearest Neighbors, Random Forests, AdaBoost, Decision Tree, Gaussian Naive Bayes, Template matching, and small NNs, were investigated. Given the small size of the dataset, a stratified K-fold cross-validation technique~\cite{james2013introduction} was employed to partition the training data into five distinct training and validation compositions.
During training, the extracted MFCCs served as input features, and the corresponding class label was used as the ground truth. To determine optimal model performance, all methods underwent a hyperparameter search across the K-folds.
Based on the evaluation metrics of accuracy and model size, the small-sized NN model shown in Figure~\ref{fig:terrain_recognition_nn_diagram} emerged as the preferred choice. This model employed a feedforward architecture with two hidden layers, each consisting of 16 neurons and utilizing Rectified Linear Unit (ReLU) activation. Additionally, batch normalization was applied to each layer to enhance the model's training. The final output layer utilized softmax activation, facilitating the generation of class probabilities.

\subsection{Implementation Details}
The chosen models for force estimation and terrain classification were designed to accommodate the constrained computational and storage capacities of Nicla Vision. The board is equipped with a powerful microcontroller (MCU) running at 400 MHz and has 1 MB of RAM and 2 MB of flash memory. The MCU present was set up using MicroPython and OpenMV firmware \cite{openmvopensource}, which supports TensorFlow Lite (TFLite) Micro~\cite{david2021tensorflow}. However, when using Micropython, the available RAM is constrained to 183 kB, imposing restrictions on the model sizes and parameters that can be stored. The NNs for both tasks were developed with Keras and TensorFlow, simplifying the process of porting them to TensorFlow Lite. The porting procedure onto the MCU avoided the use of integer quantization for weights and activations to prevent any potential loss in accuracy during the model's transfer onto the MCU. 

In order to run the force estimation model, captured images from Nicla Vision were resized using the built-in nearest neighbor method to match the required dimensions of $45\times 30$, expected by the model. On the other hand, for the terrain recognition model, a customized implementation for the MFCC extraction function was developed from the ground up and optimized using matrix-based operations. Section~\ref{sec:experiments} provides detailed insights into the performance of the implemented techniques onto the MCU.

%% file: 040_DataCollection.tex
The process to collect the data necessary to train the described machine learning methods is presented below. 

\subsection{Data Collection for Force Estimation}
\begin{figure}
    \centering
    \includegraphics[width = 0.49\textwidth]{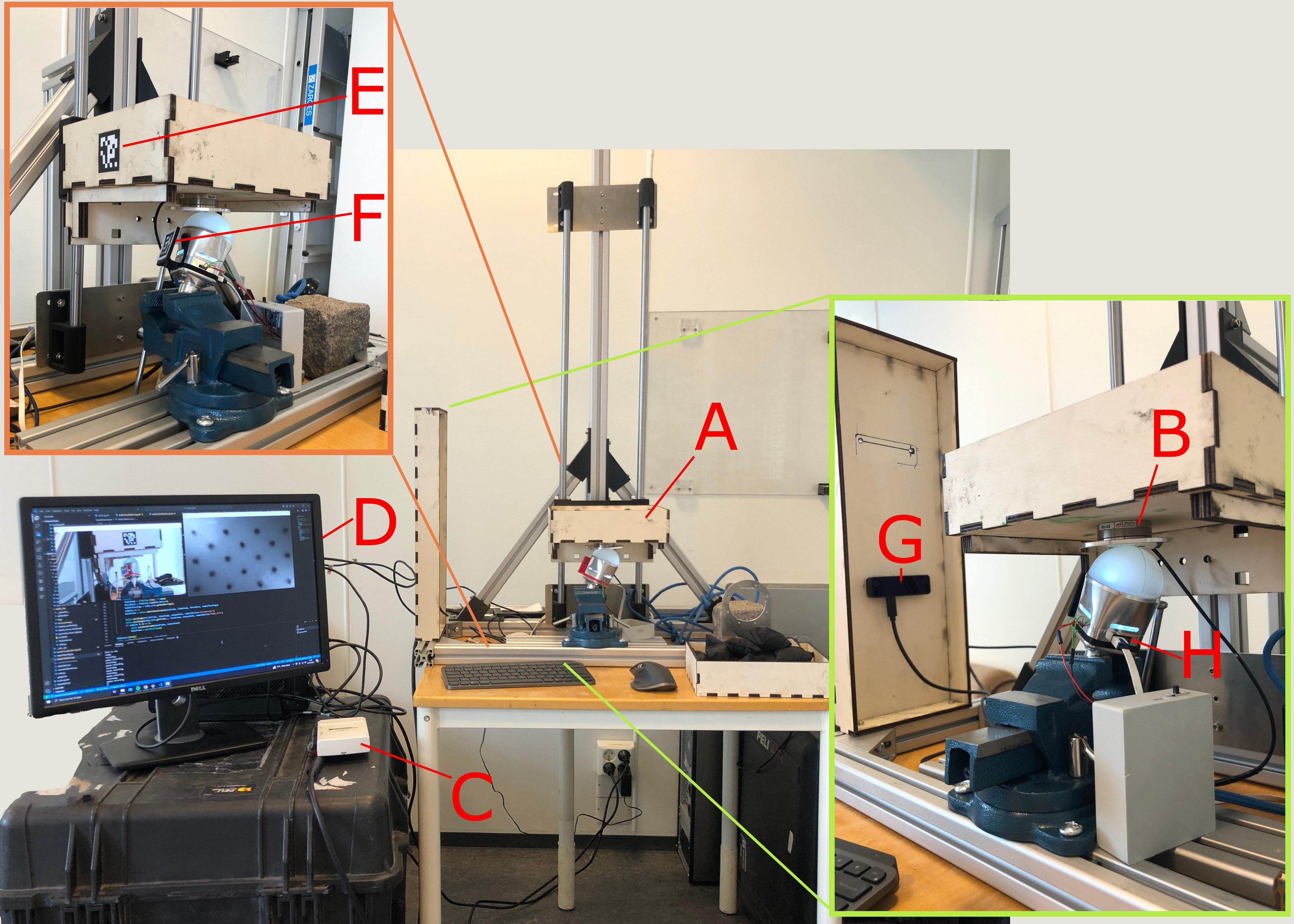}
    \vspace{-2ex}
    \caption{An overview of the experimental setup employed for data collection for the force prediction with corresponding labels assigned to each major component; A - Weight Platform, B - ATI Mini45 F/T sensor, C - data acquisition system (DAQ), D - System Computer, E - Aruco marker on the platform, F - Aruco marker on the paw, G - OAK-D Lite Camera, H - Nicla Vision inside the paw.}
    \label{fig:testrig}
\end{figure}

As mentioned in Section \ref{sec:sys_des}, in order to generate reliable ground truth data for force estimation learning, a three-part rig was designed for our experimental setup. A detailed labeled overview of the setup is illustrated in Figure~\ref{fig:testrig}. 

The diagram presents the test rig consisting of a box-shaped platform (labeled A) securely mounted on a linear rail. The platform served as a stable weight-bearing surface, effectively transferring the force to the ATI Mini45 F/T sensor (labeled B) positioned beneath it. The transducer is attached to a flat disk, ensuring a good contact surface between the transducer and the paw, facilitating accurate measurement of contact forces for ground truth. It is interfaced with a data acquisition system (DAQ) (labeled C), which is connected to a PC (labeled D) via a USB port. 

Additionally, two Aruco markers (labeled E and F) are attached to the platform and the paw, respectively. An OAK-D Lite camera (labeled G), positioned appropriately, captures the images of these markers to send to the recording PC for further processing. The detection of Aruco markers helps compute the rotation matrix necessary to determine the orientation of the applied force in relation to the paw's reference frame.

Finally, the Nicla Vision inside the paw (labeled H) completed the setup by streaming resized images to the connected PC through the USB port. To ensure synchronized data collection from the F/T transducer, OAK-D Lite camera, and Nicla Vision, a Python script was utilized, enabling simultaneous data capture from all three sensors.

Using the aforementioned setup, we collected a dataset of $17,975$ corresponding force and image data samples. The ground truth labels in Table \ref{tab:forceGroundTruth} provide information about the range of force. Figure \ref{fig:forceSamples} showcases a few examples from the dataset, illustrating the relationship between force and image data.

\begin{table}
  \centering
    \begin{tabular}{l|c|c|c}
    \hline
      \textbf{Quantity} & \textbf{x} & \textbf{y} & \textbf{z}\\
      \hline\hline
      Force Magnitude & [-66, 75] N & [-92, 80] N & [2, 133] N\\
      \hline
    \end{tabular}
    \setlength{\belowcaptionskip}{-10pt}
    \caption{Range of ground truth forces.}
    \label{tab:forceGroundTruth}
\end{table}

\subsection{Data Collection for Terrain Recognition}

\begin{figure}
    \centering
    \includegraphics[width = 0.47\textwidth]{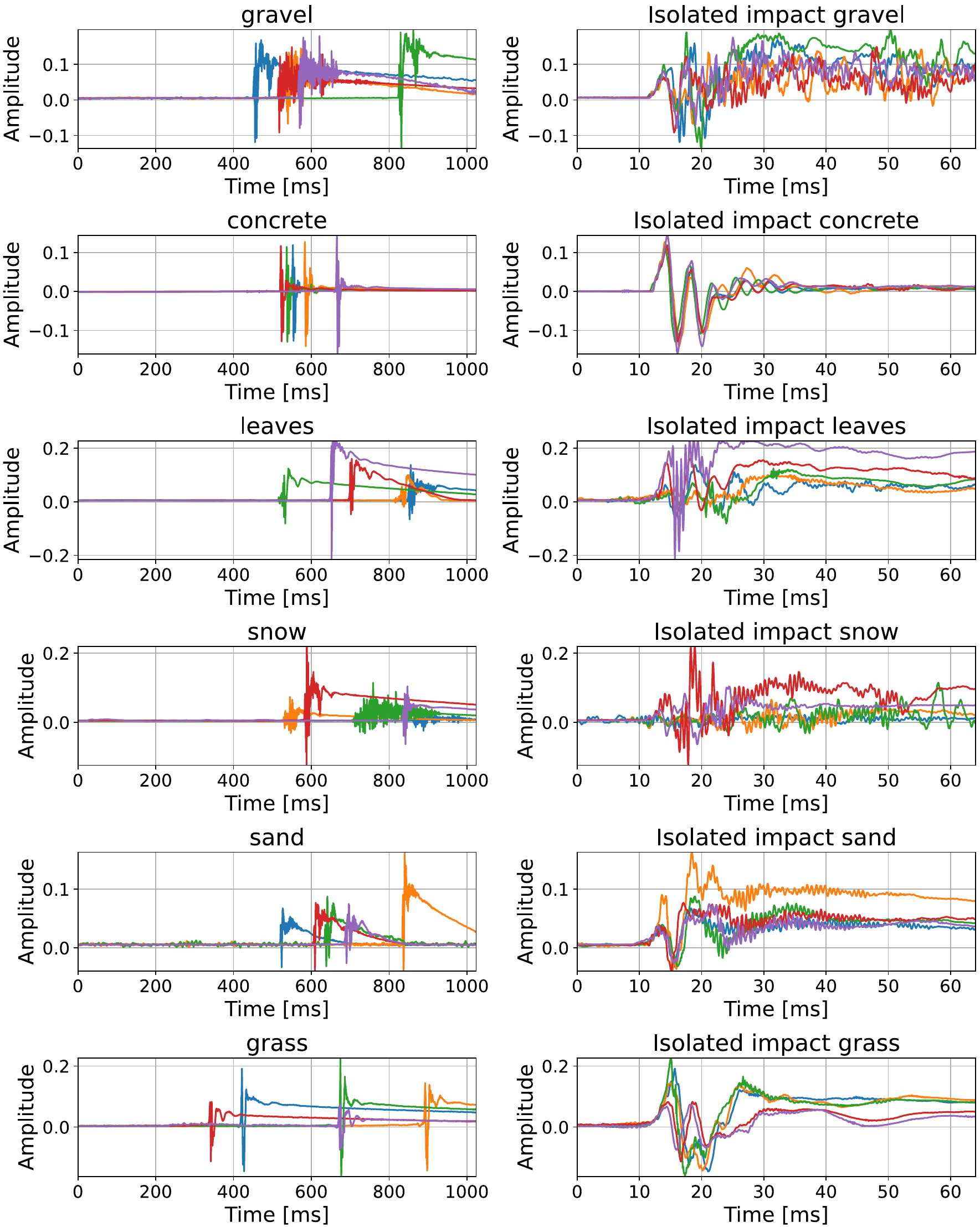}
    \setlength{\belowcaptionskip}{-10pt}
    \caption{The plots illustrate five randomly chosen raw audio signals from the complete dataset, with each plot representing a different terrain class. The left side displays the full one-second audio clips, while the right side focuses on the isolated impact sections within those clips.}
    \label{fig:TerrainAudioSamples}
\end{figure}

Alongside the image and force pairs, data were collected to train the audio-based terrain recognition neural network. For each terrain class, a total of $47$ audio samples were collected. As previously mentioned, these audio samples were truncated to a duration of $62.5$ms, removing segments with predominantly ambient noise. Figure \ref{fig:TerrainAudioSamples} showcases representative examples of the collected audio samples.

%% file: 050_Experiments.tex
A set of experimental studies serve to evaluate the performance of \ac{smartpaw} in contact force estimation and terrain recognition. 

\subsection{Force Estimation Results}
The neural network discussed in Section \ref{sec:sys_des} is trained using 80\% of the complete dataset. To select the best hyperparameters during training, 10\% of the samples are allocated as a validation set. The search results, including the evaluation of multiple models, are summarized in Table \ref{tab:ForceEstModels}. The highlighted model stands out with the lowest MAE and satisfactory inference time. The remaining 10\% of the dataset is reserved for test purposes. Figure \ref{fig:forceErrorDistribution} illustrates the error distribution for each direction, providing insight into the accuracy of the model. The resulting mean absolute error (MAE) for the applied normalized force along each direction is as follows: 0.013 N for $F_x$, 0.011 N for $F_y$, and 0.016 N for $F_z$. In order to observe the model's capability to capture the variations in the force estimation, Figure \ref{fig:forceTimeSeries} provides a visual representation of the model's performance on the test dataset. Additionally, the MAE for the total magnitude of the non-normalized forces is approximately 1.944 N, with a standard deviation of 2.711 N. The presented figure~\ref{fig:forceSamples} showcases the neural network's accurate prediction of significant force magnitude changes despite minimal visual deformation.
It highlights the network's ability to effectively capture the relationship between subtle visual cues and substantial variations in applied force. This capability can be attributed to the dataset's diversity, encompassing roll configurations from -50 to 50 degrees, pitch angle orientations from -40 to 40 degrees, and yaw angles from -180 to 180 degrees.
The chosen model has a size of 24,755 parameters, which is about 20\% of the Nicla Visions memory, leaving memory overhead for terrain classification. 


\begin{table}
    \centering
    \begin{tabular}{c|l|c|c|c}
    \hline
    Model& Structure & Parameters & MAE & Inference [$\mu$s]\\
    \hline\hline
    1& 60x90, 4, 128, 128 & 40,138 & 0.0227 & 1930 \\
    \hline
    2& 60x90, 4, 64, 64 & 26,807 & 0.0207 & 1663\\
    \hline
    3& 30x45, 8, 64 & 11,867 & 0.0206 & 630 \\
    \hline
    4& 30x45, 4, 128, 128 & 23,983 & 0.0163 & 883 \\
    \hline
    5& 30x45, 8, 256 & 14,939 & 0.0162 & 688 \\
    \hline
    6& 30x45, 8, 64, 64 & 16,283 & 0.0154 & 735 \\
    \hline
    7& 30x45, 16, 32, 32 & 23,635 & 0.0152 & 892\\
    \hline
    \textbf{8}& \textbf{30x45, 16, 128} & \textbf{24.755} & \textbf{0.0147} & \textbf{901} \\
    \hline
    \end{tabular}
    \setlength{\belowcaptionskip}{-10pt}
    \caption{Overview of the tested force prediction models, including their structure, parameter size, validation performance, and inference speed [$\mu$s] on the Nicla Vision. The last entry model is finally selected as the solution of choice for force estimation.}
    \label{tab:ForceEstModels}
\end{table}

\begin{figure}
    \centering
    \includegraphics[width = 0.4\textwidth]{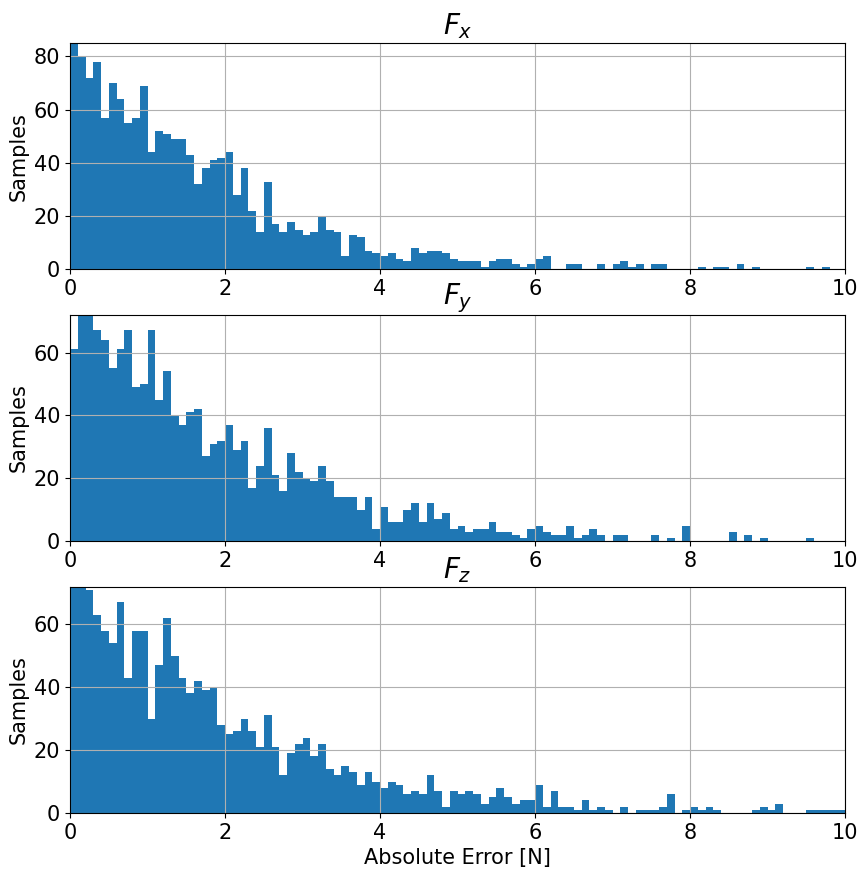}
    \setlength{\belowcaptionskip}{-10pt}
    \caption{Histogram displaying the mean absolute error (MAE) between the force prediction and the corresponding ground truth values for the test dataset.}
    \label{fig:forceErrorDistribution}
\end{figure}

\begin{figure}
    \centering
    \includegraphics[width = 0.4\textwidth]{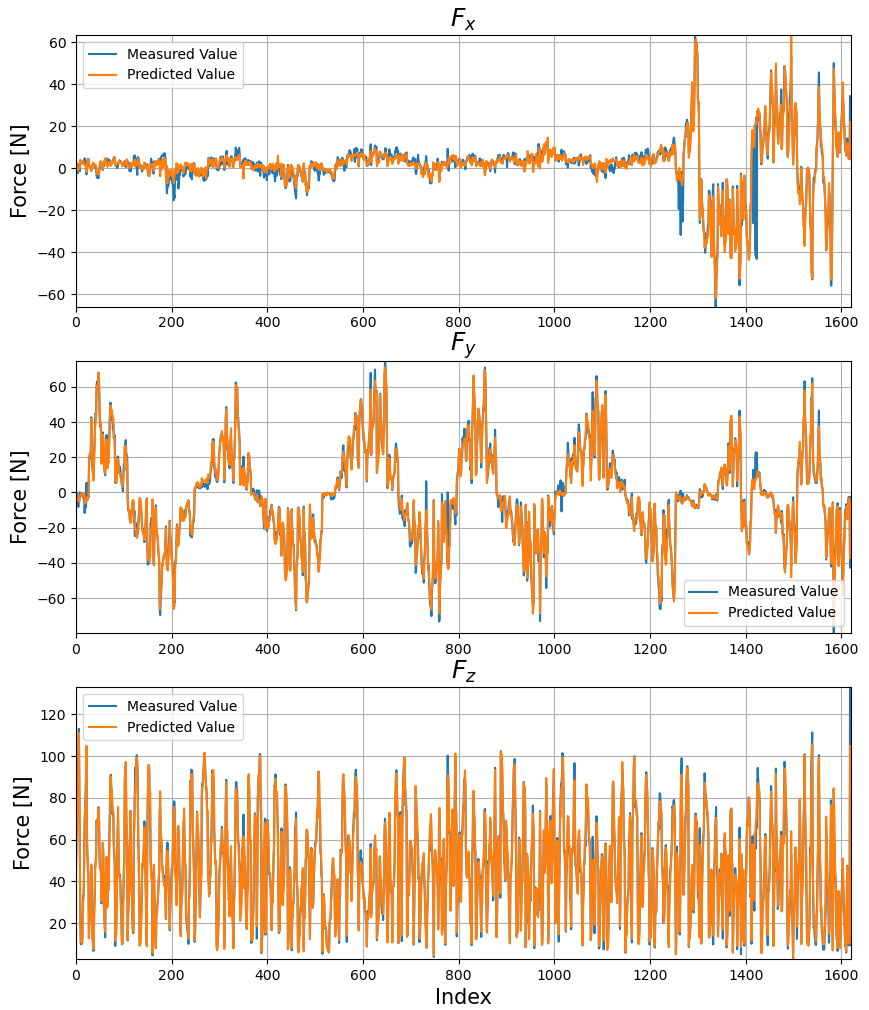}
    \setlength{\belowcaptionskip}{-10pt}
    \caption{Timeseries comparing the force predictions (in orange) and the ground truth values (in blue) across all three axes for the test dataset.}
    \label{fig:forceTimeSeries}
\end{figure}

\begin{figure}
  \centering
  \begin{subfigure}[t]{.23\textwidth}
    \centering
    \includegraphics[width=\linewidth]{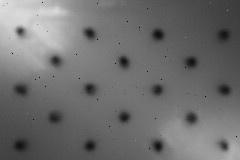}
    \caption{Magnitudes(N): 2.84, 2.41}
  \end{subfigure}
  \hfill
  \begin{subfigure}[t]{.23\textwidth}
    \centering
    \includegraphics[width=\linewidth]{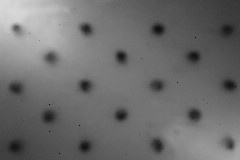}
    \caption{Magnitudes(N): 45.06, 43.56}
   \end{subfigure} 
  \medskip

  \begin{subfigure}[t]{.23\textwidth}
    \centering
    \includegraphics[width=\linewidth]{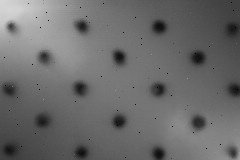}
    \caption{Magnitudes(N): 90.54, 88.62}
  \end{subfigure}
  \hfill
  \hspace{-30pt}
  \begin{subfigure}[t]{.23\textwidth}
    \centering
    \includegraphics[width=\linewidth]{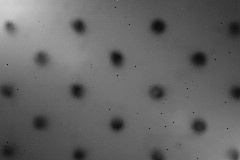}
    \caption{Magnitudes(N): 134.41, 106.79}
  \end{subfigure}
  \setlength{\belowcaptionskip}{-10pt}
  \caption{This figure displays four instances where the neural network made predictions on the provided images. The captions for each image present the predicted force magnitudes obtained from ground truth data (left value) and the corresponding force magnitudes computed by the neural network (right value).}
  \label{fig:forceSamples}
  \vspace{-2ex}
\end{figure}

\subsection{Terrain Classification Results}
The terrain classification methods discussed in Section \ref{sec:sys_des} were trained on a subset of the dataset, constituting 80\% of the total samples. To extend the usage of the dataset, a K-fold technique was employed to evaluate the performance of the models, which involved dividing the learning dataset into 80\% training and 15\% validation subsets. The experimented models, along with their respective sizes and cross-validation scores, are listed in Table \ref{tab:TerrainClassificaitonModels}. 

Here it can be seen that sparse models such as Gaussian Naive Bayes (GNB), Template matching, and a small NN provide the best accuracy and standard deviation regardless of their size. Other methods such as Support Vector Machines (SVM) and k-Nearest Neighbors (k-NN) present worse performance. The selected NN model has a cross-validation mean of $0.779$ and a standard deviation of $0.039$ while keeping a small size of $238$ parameters. 
To ensure that the NN, which had the best accuracy, did not suffer from overfitting, it was tested using the test dataset, as seen in Figure \ref{fig:TerrainConfusionMatrix}. The overall performance is high, while the gravel, snow, and sand classes are the least distinguishable.

The terrain classification based on MFCCs of the impact sounds was evaluated using nine different ML methods. Also, the models were evaluated based on their cross-validation accuracy and size, revealing the best-performing method being a small NN.

\begin{table}[!htbp]
    \centering
    \begin{tabular}{c|c|c|c}
        \hline
        Model Name & Size & Cross Val. Mean & Cross Val. std\\
        \hline\hline
        SVM & 17.527 & 0.718 & 0.106 \\
        \hline
        k-NN & 29.108 & 0.629 & 0.112\\
        \hline
        Random Forest & 74.490 & 0.667 & 0.088 \\
        \hline
        AdaBoost & 88.229 & 0.665 & 0.0123 \\
        \hline
        Decision Tree & 7.463 & 0.486 & 0.200 \\
        \hline
        GNB & 1.936 & 0.736 & 0.127 \\
        \hline
        QDA & 10.350 & 0.314 & 0.094  \\
        \hline
        Template Matching & 1.009 & 0.681 & 0.052 \\
        \hline
        \textbf{Neural Network} & \textbf{3.128} & \textbf{0.779} & \textbf{0.039} \\
        \hline
    \end{tabular}
    \setlength{\belowcaptionskip}{-15pt}
    \caption{Overview of the tested model types for terrain classification, their parameter size, cross-validation mean, and cross-validation standard deviation. The last entry model, using a neural network, is finally selected as the solution of choice for terrain classification.}
    \label{tab:TerrainClassificaitonModels}
\end{table}

\begin{figure}
    \centering
    \includegraphics[width=0.48\textwidth]{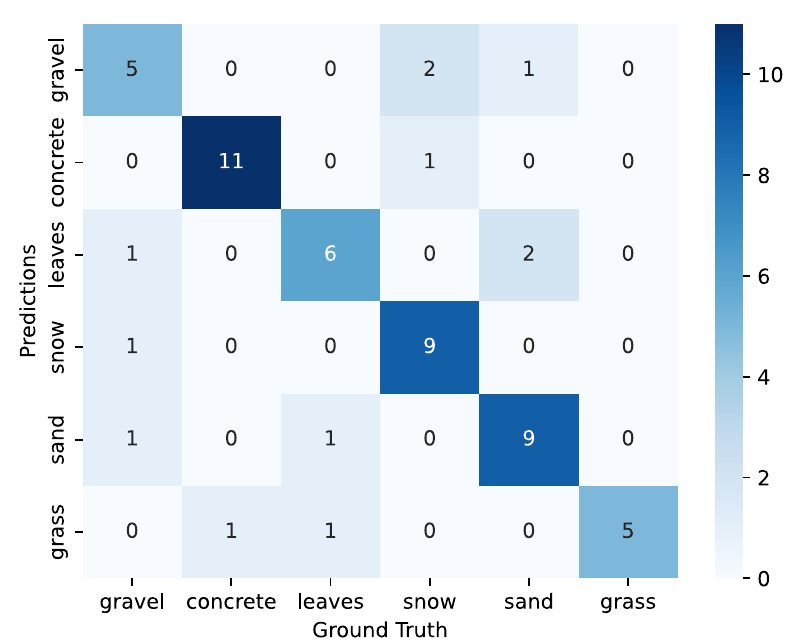}
    \setlength{\belowcaptionskip}{-15pt}
    \caption{Confusion matrix depicting the prediction summary achieved by the selected neural network for the six distinct terrain classes.}
    \label{fig:TerrainConfusionMatrix}
\end{figure}

\subsection{Computational Metrics}
We conducted a comprehensive evaluation of performance, efficiency, and resource utilization for both functionalities implemented on the MCU.

For the force prediction task, which involved image capturing, resizing, and model inference, we timed the completion of $10,000$ passes. Each pass averaged approximately $0.9$ milliseconds, making it $25\times$ faster than the camera frame rate of $43$ FPS. Memory consumption during execution was approximately $34.5$ KB of RAM. The combination of compact size, high accuracy, and efficient runtime established this method as an effective solution for contact force estimation.

Similarly, for terrain recognition, we assessed the MFCC extraction process and subsequent TFLite model inference using pre-recorded audio samples. Conducting $10,000$ passes on the $1$-second samples, MFCC extraction averaged around $15.2$ milliseconds per sample, while TFLite model inference required approximately $0.13$ milliseconds per sample. Memory consumption was measured at about 50 KB of RAM for both processes, taking up $25\%$ of its available memory.

%% file: 060_Conclusion.tex
In this paper, the \ac{smartpaw} system has been introduced, and its effectiveness in accurately estimating ground contact forces and delivering terrain classification has been demonstrated through experimental tests. The system incorporates sensing and computing on edge, with the neural network inference run times for force estimation and terrain recognition being sufficiently fast for real-time use onboard the Nicla Vision. As future work, certain improvements are possible. First, the data collected for training force prediction can be enhanced by recordings on uneven terrain. Likewise, regarding the audio-based terrain recognition solution, although the overall performance is high, it must be noted that the presented terrain classes presented high distinguishability, and the method would have to be further assessed in more complex multi-classed terrain.